\documentclass[10pt,twocolumn,letterpaper]{article}

\usepackage{cvpr}

\usepackage{graphicx}
\usepackage{amsmath}
\usepackage{amssymb}
\usepackage{booktabs}

\usepackage{color}

\usepackage{epstopdf}
\usepackage{multirow}
\usepackage{amsmath,amssymb}
\usepackage{algorithm}
\usepackage{algorithmicx}
\usepackage{algpseudocode}

\usepackage{booktabs}
\usepackage{makecell}
\usepackage{textcomp}
\usepackage{mathrsfs}
\usepackage{multicol}
\usepackage{mdframed}
\usepackage[T1]{fontenc}

\usepackage[pagebackref,breaklinks,colorlinks]{hyperref}

\usepackage[capitalize]{cleveref}
\crefname{section}{Sec.}{Secs.}
\Crefname{section}{Section}{Sections}
\Crefname{table}{Table}{Tables}
\crefname{table}{Tab.}{Tabs.}

\begin{document}

\title{EMOdiffhead: Continuously Emotional Control in Talking Head Generation via Diffusion}

\author{Jian Zhang  \\
South China University of Technology\\
{\tt\small 202221018425@mail.scut.edu.cn}
 \and
Weijian Mai \\
South China University of Technology\\
{\tt\small 202221018451@mail.scut.edu.cn}
 \and
Zhijun Zhang $*$  \\
South China University of Technology\\
{\tt\small auzjzhang@scut.edu.cn}
 \\
}
\maketitle








\begin{abstract}
The task of audio-driven portrait animation involves generating a talking head video using an identity image and an audio track of speech. While many existing approaches focus on lip synchronization and video quality, few tackle the challenge of generating emotion-driven talking head videos. The ability to control and edit emotions is essential for producing expressive and realistic animations. In response to this challenge, we propose EMOdiffhead, a novel method for emotional talking head video generation that not only enables fine-grained control of emotion categories and intensities but also enables one-shot generation. Given the FLAME 3D model's linearity in expression modeling, we utilize the DECA method to extract expression vectors, that are combined with audio to guide a diffusion model in generating videos with precise lip synchronization and rich emotional expressiveness. This approach not only enables the learning of rich facial information from emotion-irrelevant data
but also facilitates the generation of emotional videos. It effectively overcomes the limitations of emotional data, such as the lack of diversity in facial and background information, and addresses the absence of emotional details in emotion-irrelevant data. Extensive experiments and user studies demonstrate that our approach achieves state-of-the-art performance compared to other emotion portrait animation methods.
\end{abstract}

\section{Introduction}\label{introduction}
The task of audio-driven portrait animation involves creating a talking head video using an identity image of the speaker and an audio track of their speech content, that has a wide range of applications. For example, it can serve as a means of human-computer interaction and is widely used in digital assistants, film-making, and virtual video conferences.

With the development of AI-generated content, especially generative adversarial networks (GAN) and diffusion models, talking head generation has become a research hotspot in recent years due to its profound application prospects. Most of the previous works focused on solving lip synchronization and video quality issues, and only a few works explored generating emotion-related videos. However, the task of emotion-driven talking head generation is a key aspect in producing animated faces that are not only realistic but also capable of conveying a wide range of emotions with remarkable expressiveness. Although some talking head generation studies focus on incorporating facial emotions, they are unable to flexibly edit these expressions, such as adjusting the intensity of the expression.

\begin{figure}[htbp]
\centering
\includegraphics[scale=0.40]{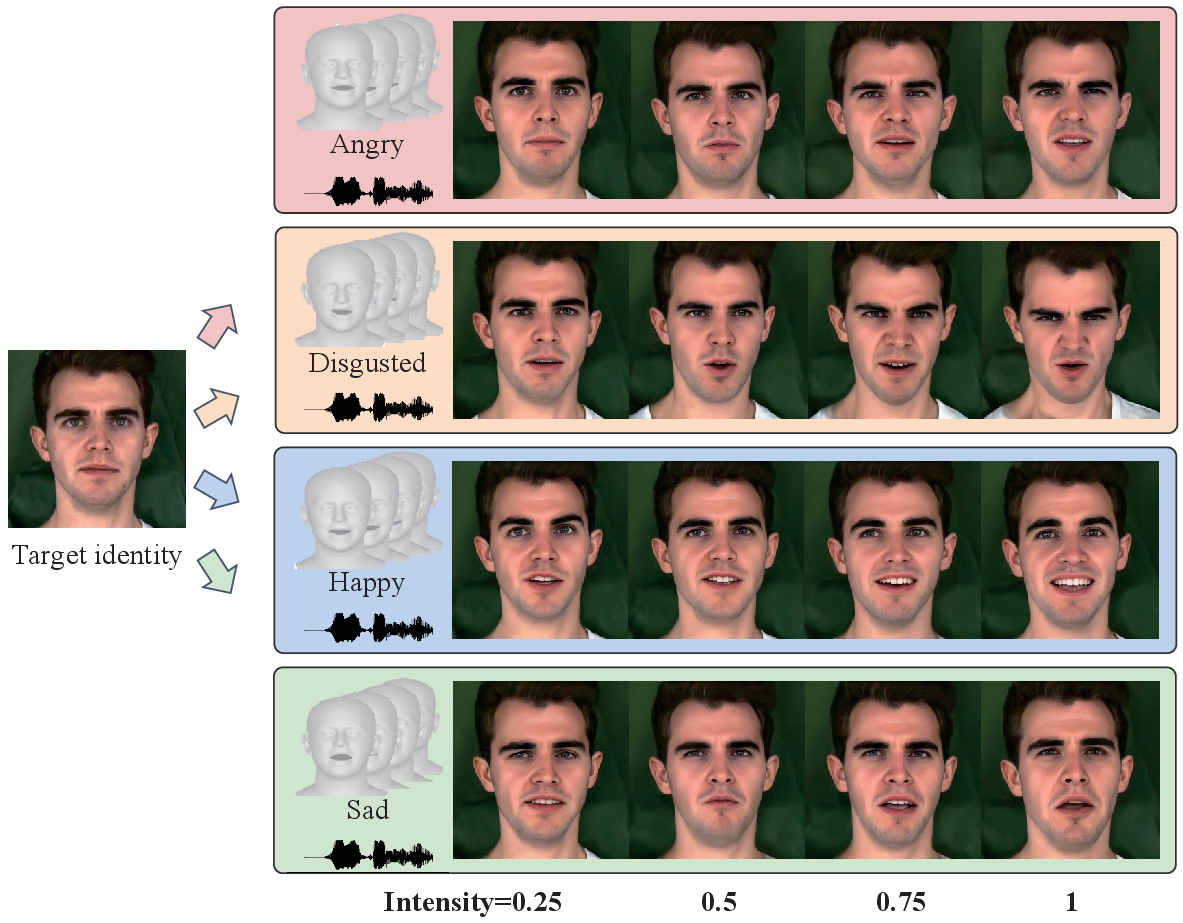}
\centering
\caption{Given an emotion label and specified intensity, our method first generates an expression vector. This vector is then combined with the audio and the target identity image to synthesize a video that aligns with the specified emotion and intensity.}\label{Rendering}
\end{figure}

When it comes to emotion-driven talking head generation, early works such as \cite{eskimez2021speech} and \cite{sinha2022emotion} use one-hot emotion label as the emotion source, without the capability of intensity editing. EVP \cite{ji2021audio} presents a disentanglement method to separate emotion latent feature representation from speech for implicitly controlling the emotion of synthesized video. However, the effective representation of latent emotions in EVP depends on the accuracy of the speech content and emotion disentanglement method, thus it is difficult to achieve emotion control entirely relying on speech alone. Moreover, it is also inconvenient if the input emotion source depends on other videos \cite{ji2022eamm}, since finding a video with the desired emotional style may not always be feasible. For instance, finding the desired emotional video requires considering factors such as resolution, occlusion, and even the length relationship between the video and audio. In addition, large-scale emotion-irrelevant audio-visual datasets are more easily available than emotional data. These data are typically collected in the wild, therefore containing diverse facial and background information, that is beneficial to enhance the model's generalization performance. Therefore, there exist two key issues that need to be addressed. \textbf{(1) How to effectively and flexibly control the emotions of synthesized videos?} \textbf{(2) How to effectively learn emotional information from emotion-irrelevant data to improve generalization performance?}

To address the challenge mentioned above, we propose the method EMOdiffhead that can explicitly and continuously edit the emotion of talking head video, achieving fine-grained control as shown in Figure \ref{Rendering}. Specifically, given that the FLAME 3D model allows for linear editing in expression modeling, we employ the DECA method to reconstruct the facial geometry from the video. It enables us to extract the expression vector and create a training pair consisting of the video’s emotion label and the corresponding expression vector. Subsequently, given the input audio and expression vector, a time-based denoising network is trained to generate a video matching the input. For maintaining the consistency of the target subject, ReferenceNet is introduced to extract the feature map of the target identity that is to be integrated into the denoising network through cross-attention. To synthesize videos that match emotions and intensities during inference, the above emotion label and expression vector pairs are used to train an expression vector generator. Next, the synthesized neutral expression vector and the target emotional expression vector generated by the generator are used to compute the final expression input vector. It is then used to condition the denoising network to synthesize videos that correspond to the desired emotions and their intensities.

Our principal contributions are as follows:

\begin{itemize}
 \item We propose EMOdiffhead, an innovative method for emotional talking head generation with fine-grained control and one-shot generation capabilities. It uniquely leverages the FLAME 3D model's emotion encoding vector as the condition, allowing for flexible control of emotion categories and intensities while ensuring precise lip synchronization.

 \item We propose a new metric named FLIE, to evaluate the linearity of emotion intensity editing in talking head generation.

 \item To the best of our knowledge, our method can effectively learn emotional information from emotion-irrelevant data. By employing the DECA method to extract expression vectors as a condition for generating emotional faces, it can not only learn rich facial information from emotion-irrelevant data but also achieve emotional video generation.
\end{itemize}

\section{Related Work}\label{sec.related_work}
Recently, researchers have proposed various methods for audio-driven talking head synthesis using deep neural networks. In the following, we discuss prior works in audio-driven talking head generation and condition emotion generation.
\subsection{Audio-driven talking head generation}

Audio-driven talking head generation is a technique that employs audio to generate facial animations or expressions that correspond to the audio content. It can be divided into two types, i.e., \textbf{Person-specific} and \textbf{Person-independent}. Person-specific talking face generation \cite{guo2021ad,ye2023geneface,zhang2021facial,sun2021speech2talking} have the advantage of producing high-resolution talking face videos since the identity is included in the training data. However, a drawback of the Person-specific approach is that the model is difficult to generalize to other identities. Therefore, Person-independent methods have emerged to solve this problem. Person-independent approaches that can generalize to arbitrary faces are trained on large-scale audio-visual datasets with diverse faces, lighting, and backgrounds, such as Voxceleb \cite{nagrani2017voxceleb}. In previous works, Chung et al. \cite{chung2017you} first generate talking faces in a one-shot manner. Chen et al. \cite{chen2019hierarchical} and Zhou et al. \cite{zhou2020makelttalk} improve on this scheme by leveraging facial landmarks as intermediate representations. Zhou et al. \cite{zhou2021pose} further incorporate pose control into the one-shot setting. Zhang et al. \cite{zhang2023sadtalker} and Ren et al.\cite{ren2021pirenderer} use 3D coefficient representation for one-shot generation. Despite these advancements, these methods still struggle to generate animations with varying emotional expressions.

\subsection{Condition Emotion Generation}
Recently, the focus has expanded to controlling the emotions of the output subject. Because emotions play a vital role in realistic animation. Eskimez et al. \cite{eskimez2021speech} and Sinha et al. \cite{sinha2022emotion} employ a one-hot emotion label to generate an emotional talking face but the method is unable to control the intensity of emotion. EVP \cite{ji2021audio} presents a disentanglement approach to decouple content and emotion information from speech for implicitly controlling the emotion of output video. However, the method cannot be applied to unseen subjects and audio. Furthermore, these methods for generating emotional talking faces trained on the emotional audio-visual datasets CREMA-D \cite{cao2014crema} and MEAD \cite{wang2020mead} have limited generalization capability due to the low diversity of these datasets. For the one-shot method, EAMM \cite{ji2022eamm} achieves accurate emotional control of synthesized videos by using emotional features extracted from reference videos, but it is not flexible. Although Tan et al. \cite{tan2024style2talker} use user-friendly text descriptions to specify the desired emotional style, they cannot achieve fine-grained control. In contrast, our proposed method EMOdiffhead can achieve fine-grained emotion control and remains effective for arbitrary identities.

\begin{figure*}[htbp]
\centering
\includegraphics[scale=0.7]{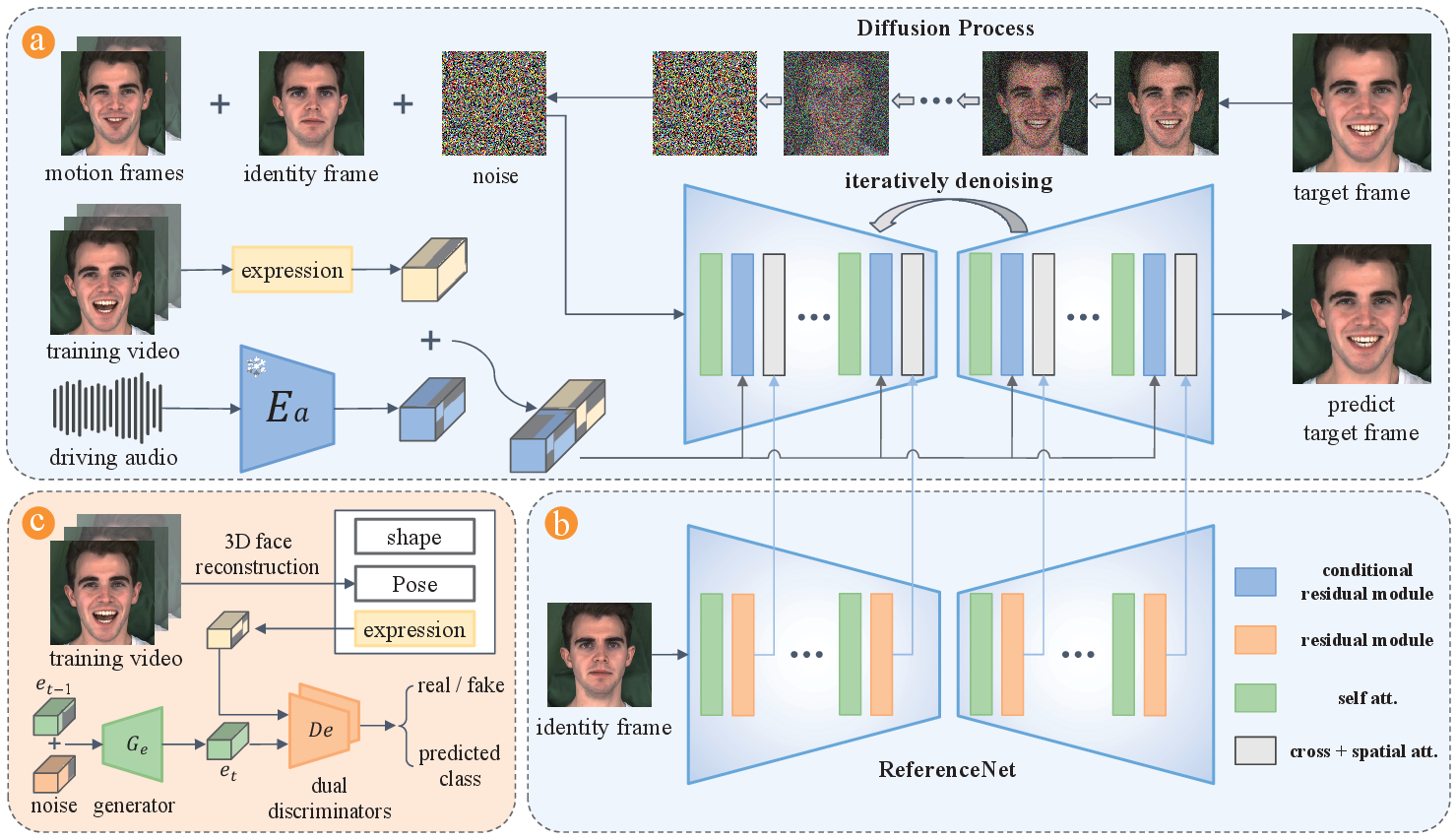}
\centering
\caption{Pipeline of our method. Given an emotion label with a specified intensity, our method first generates FLAME-based expression vectors. Next, these vectors are then combined with a reference image and audio to condition the diffusion model for generating the target video. \textbf{(a) Backbone Network:} The time-based diffusion model receives image, audio, and expression vector input to synthesize the target video. \textbf{(b) ReferenceNet:} Another UNet with a similar structure to the backbone network is used to extract features of a reference image for maintaining identity consistency.
\textbf{(c) Emotion Editing Condition Generation Module:} Manually specify emotion category and intensity. During inference, the editing direction between the target non-neutral emotion and the neutral emotion and intensity value are utilized to obtain the final expression vectors.}\label{fig:overview}
\end{figure*}

\section{Method}\label{sec.method}
Given a target identity image, an audio clip, and an emotion label with intensity, our proposed method, EMOdiffhead, can generate talking head videos that are synchronized with the audio, reflect the specified emotional intensity, and remain consistent with the target identity. The EMOdiffhead framework consists of two main modules: emotion editing condition generation and photo-realistic talking head synthesis. An outline of the proposed pipeline is presented in Figure \ref{fig:overview}.

\subsection{3D Face Reconstruction for Extracting Expression Vectors}\label{FaceReconstruction}
To achieve continuous linear control of expressions in talking head video synthesis, we leverage the 3D Morphable Model (3DMM) to estimate facial expression parameters, providing an effective representation of emotions. Specifically, we perform 3D face reconstruction using the DECA method \cite{feng2021learning}, which employs the FLAME model \cite{li2017learning}. FLAME is a 3D facial statistical model that integrates linear identity shape and expression modeling. For each frame in talking head video, the DECA method can regress the FLAME parameters of human face, including the parameters of identity shape $\boldsymbol{\beta}\in\mathbb{R}^{100}$, expression $\boldsymbol{\psi}\in\mathbb{R}^{50}$ and head pose $\boldsymbol{\theta}\in\mathbb{R}^{6}$, etc. Subsequently, the expression vector $\boldsymbol{\psi}$ from the training video is then used as an additional emotional condition, embedded into the process of talking head synthesis.

\subsection{Photo-realistic Talking Head Synthesis}
To tackle the challenge of generation quality and model
generalization, diffusion models with strong generative capabilities are utilized to synthesize talking head videos. To synchronize the generated video with the driving audio and match the corresponding intensity of emotions, we inject additional audio embeddings and emotion embeddings during the progressive denoising process. Additionally, we employ a UNet architecture, similar to the denoising backbone network but without temporal information, to extract identity features. These features are then used to perform cross-attention calculations with the corresponding layers in the backbone network, ensuring identity consistency in the generated video. Finally, the output video undergoes super-resolution module processing to obtain high-resolution photo-realistic talking head video.

\subsubsection{Diffusion Models}
Diffusion models involve two key processes, i.e., diffusion and denoising. The diffusion process progressively adds noise to the data, effectively destroying its structure, while the denoising process learns to reverse this by restoring the data. In practice, only the denoising process is trained and used during the inference stage, where it generates data by iteratively refining a sample of Gaussian noise through multiple denoising steps.

In the diffusion process, given samples $x_0$ from distribution $P_\text{data}(x_0)$, noise is gradually added over a series of time steps $t=1,2,…,T$. At each step $t$, a small amount of Gaussian noise is introduced, creating a sequence of increasingly noisy samples $x_1,x_2,…,x_T$. This process is governed by the following Gaussian transition:

\begin{align}
q(x_t|x_{t-1})=\mathcal{N}(x_t;\sqrt{1-\beta_t}x_{t-1},\beta_t \textbf{I})\label{formula1}
\end{align}
where $\beta_t$ is a variance schedule that typically increases over time, and $\mathcal{N}$ represents the Gaussian distribution. As $t$ increases, the samples $x_t$ become progressively noisier, eventually approximating pure Gaussian noise when $t$ is large.

Finally, sample $x_t$ at any step $t$ is allowed for the direct computation from the original data point $x_0$. This can be expressed as:

\begin{align}
x_t = \sqrt{\bar{\alpha}_t}x_0 + \sqrt{{1 - \bar{\alpha}_t}}\epsilon\label{formula2}
\end{align}
where $\alpha_{t}=1-\beta_{t}$, $\bar{\alpha}_t=\prod_{s=1}^{t} a_s$, and $\epsilon \sim \mathcal{N}(0, \textbf{I})$ is Gaussian noise.

The reverse process is the generative phase where we start from a sample of pure Gaussian noise $x_T$ and iteratively remove noise to recover a sample that follows the data distribution $P_\text{data}(x_0)$. This reverse process can be modeled as a Markov chain with learned parameters. Each step of the reverse process is defined as follows:
\begin{align}
p_\theta(x_{t-1}\mid{x_t}) = \mathcal{N}(x_{t-1}; \mu_\theta(x_t, t), \Sigma_\theta(x_t, t))\label{formula3}
\end{align}

As described by [23], $\mu_\theta(x_t, t)$ and $\Sigma_\theta(x_t, t)$ are further reparameterized into:
\begin{align}
\mu_\theta(x_t, t)=\frac{1}{\sqrt{\alpha_{t}}}(x_t-\frac{\beta_{t}}{\sqrt{1-\bar\alpha_{t}}}\epsilon_\theta(x_t,t))\label{formula4}\\
\Sigma_\theta(x_t, t)=\text{exp}(v\text{log}\beta_{t}+(1-v)\text{log}\tilde\beta_{t})\label{formula5}
\end{align}
where $\tilde\beta_{t}=\frac{1-\bar\alpha_{t-1}}{1-\bar\alpha_{t}}\beta_{t}$, $\epsilon_\theta(x_t,t)$ and $v$ is the predict result of neural networks. The $\epsilon_\theta(x_t,t)$ and $v$ are trained using $\mathcal{L}_\text{simple}$ and $\mathcal{L}_\text{vlb}$ respectively, where:
\begin{align}
\mathcal{L}_\text{simple} &= \text{E}_{t,x_0,\epsilon}[\|\epsilon-\epsilon_\theta(x_t,t)\|^2]\label{formula6}\\
\mathcal{L}_\text{vlb} &= \mathcal{L}_0+\mathcal{L}_1+...+\mathcal{L}_{T-1}+\mathcal{L}_{T}\label{formula7}\\
\mathcal{L}_0 &= -\text{log}p_{\theta}(x_0|x_1)\label{formula8}\\
\mathcal{L}_{t-1} &= D_{KL}(q(x_{t-1}|x_t,x_0)||p_\theta(x_{t-1}|x_t))\label{formula9}\\
\mathcal{L}_T &= D_{KL}(q(x_T|x_0)||p(x_T))\label{formula10}
\end{align}

Notably, except $\mathcal{L}_0$, each term in Equation \ref{formula7} is the KL divergence of two Gaussians. And $\mathcal{L}_T$ is discarded during training because it doesn't depend on $\theta$, it will be close to 0 if the diffusion process perturbs the data adequately so that ${q(x_T|x_0)}\approx{\mathcal{N}(0,\textbf{I})}$. Generally, UNet is utilized as the backbone to predict $\epsilon_\theta(x_t,t)$ and $v$ for computing $\mu_\theta(x_t, t)$ and $\Sigma_\theta(x_t, t)$ respectively. During inference, a new sample is synthesized by sampling $x_T$ from $\mathcal{N}(0,\textbf{I})$ and iteratively denoising according to Equation \ref{formula3}.

\subsubsection{Model Architecture}
Figure \ref{fig:overview} depicts the architecture of the EMOdiffhead method. For achieving video editing, additional information such as audio, identity image, and emotion with intensity are provided for guiding its generation process. In our case, we incorporate this information into a time-conditional UNet by utilizing cross-attention blocks and conditional residual blocks. Details of the conditional mechanism in our proposed method
are elaborated below.

\textbf{Frame-based conditioning.} Given a video frame sequence $X=\{x^{(0)},x^{(1)},...,x^{(N)}\}$ of length $N$, our model takes three type images as input, i.e., noisy frame $x^{(i)}_t$, identity frame $x_\text{id}$ and motion frames $x_\text{motion}$. The noisy frame is obtained by adding noise to the target frame at the $t$ time step. The identity frame is randomly sampled from $X=\{x^{(0)},x^{(1)},...,x^{(N)}\}$. It not only maintains the consistency of the target identity but also improves the generation robustness of the model. Moreover, to smooth the generating video, for the target frame $x^{(i)}$, we introduce the motion frames ${x^{(i-2)},x^{(i-1)}}$ to provide more temporal information for the model. Finally, the three type frames are concatenated at channel dimension and fed into the backbone as input. The final input $x^{(i)}_\text{input}$ is shown as follows:
\begin{align}
x^{(i)}_\text{input}=x^{(i)}_t \oplus x_\text{id} \oplus x^{(i-2)}\oplus x^{(i-1)}\label{formula15}
\textit{}\end{align}
where $\oplus(\cdotp)$ represents concatenate operation.

\textbf{Audio Conditioning.} For a given video frame sequence $X=\{x^{(0)},x^{(1)},...,x^{(N)}\}$, we must preprocess its original audio source so that corresponds to the number of video frames. Specifically, we utilize the pretrained audio encoder to encode the original audio source so that obtain the audio feature $A=\{a^{(0)},a^{(1)},...,a^{(N)}\}$. Next, we inject this audio information into the UNet by using conditional residual blocks, specifically following the approach of \cite{stypulkowski2024diffused} to scale and shift the hidden states of the UNet:
\begin{align}
h_{s+1}=a^{(i)}_s(t_s \text{GN}(h_s)+t_b)+a^{(i)}_b\label{formula11}
\end{align}
where $h_s$ and $h_{s+1}$ are consecutive hidden states of UNet, $\text{GN}(\cdotp)$ represents group normalization, $(t_s,t_b)=\text{MLP}(t^{'})$, $(a^{(i)}_s,a^{(i)}_b)=\text{MLP}(a^{(i)'})$, $t^{'}$ and $a^{(i)'}$ are the embedding output by the embedding layer of time $t$ and audio $a^{(i)}$ respectively. $\text{MLP}(\cdotp)$ represents a shallow neural network composed of a linear layer and a SiLU activation function.

To ensure the accuracy of synthesized lip movements, we consider past and future audio segments and splice them with the audio of the current frame. Thus $a^{(i)'}$ is replaced by the audio embeddings combination $\{a^{(i-n)'},...,a^{(i)'},...,a^{(i+n)'}\}$ to participate in the calculation in Equation \ref{formula11}, where $n$ is the number of additional audio embeddings from one side.

\textbf{Emotion Conditioning.} In the training stage, we inject the emotional information corresponding to each frame so that the backbone network can synthesize videos that meet the specified emotional conditions. Although some existing datasets about talking face videos have given corresponding emotion labels and intensity levels, it is difficult to directly inject them into the backbone network as emotional conditions to achieve fine-grained control of emotions. As mentioned in Section \ref{FaceReconstruction}, the DECA method can reconstruct expression vectors from a single face image. Therefore, the expression vectors $E=\{e^{(0)},e^{(1)},...,e^{(N)}\}$ extracted by using DECA to reconstruct faces from video frame sequence $X=\{x^{(0)},x^{(1)},...,x^{(N)}\}$ can be used as effective expression representations.

Inspired by \cite{stypulkowski2024diffused}, we utilize the same condition injection method as audio to introduce expression information. This can be expressed as:

\begin{align}
h_{s+1}=e_s^{(i)}(a^{(i)}_s(t_s \text{GN}(h_s)+t_b)+a^{(i)}_b)+e_b^{(i)}\label{formula14}
\end{align}
where $(e^{(i)}_s,e^{(i)}_b)=\text{MLP}(e^{(i)'})$, and $e^{(i)'}$ is the embedding
output by the embedding layer of expression $e^{(i)}$. To make use of expression information, we use $\{e^{(i-n)'},...,e^{(i)'},...,e^{(i+n)'}\}$ instead of $e^{(i)'}$ as the expression embedding of the current video frame to participate in the calculation of Equation \ref{formula14}.

\begin{figure}[htbp]
\centering
\includegraphics[scale=0.42]{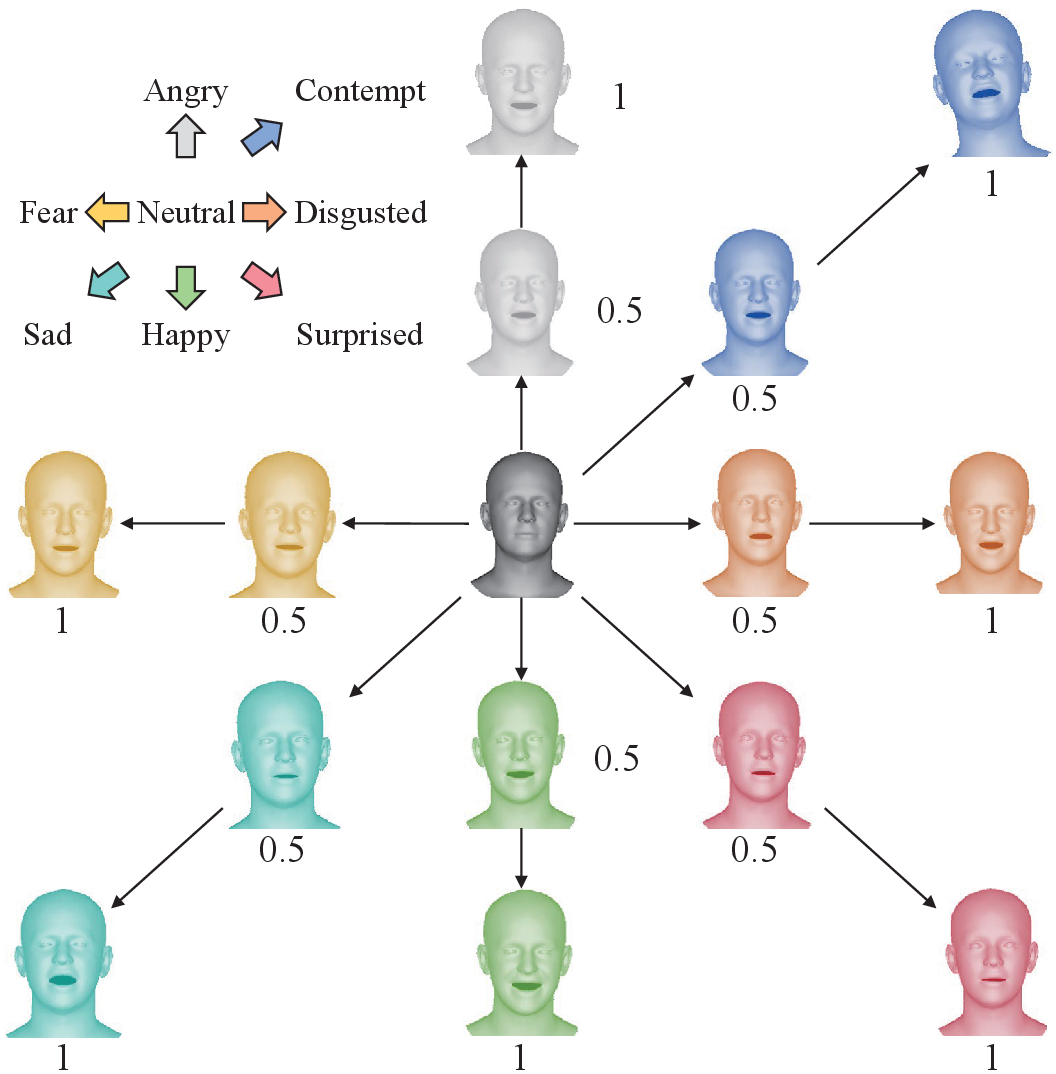}
\centering
\caption{The FLAME model's linear editing characteristics allow for gradual changes in emotion intensity: as the natural emotion vector shifts toward the unnatural emotion vector, the intensity of the unnatural emotion progressively increases. The numbers in the figure represent the strength values. }\label{expression}
\end{figure}

Since large-scale audio-visual datasets generally contain a wide range of videos with different emotion types and intensities but lack corresponding labels, using the expression vector corresponding to each video frame as a condition during training can guide the backbone network to synthesize videos with varying emotions. At the same time, due to the linear editing characteristics of the FLAME model in the expression space (shown as Figure \ref{expression}), when given a series of expression vectors that continuously change along the direction from neutral emotions to non-neutral emotions, e.g., angry, contempt, disgusted, fear, happy, sad and surprised during inference, our network can synthesize expressions continuously changing talking head videos. Specifically, given a pair of expression vectors, one corresponds to neutral emotion and the other corresponds to target non-neutral emotion, we can calculate the editing direction vector for target non-neutral emotion. When we specify the intensity of the given target emotion $y$, we can obtain the final expression vector $e_o$:
\begin{align}
e_o&=e_n+k*d_{y},  k\in[0,1]\label{formula12}\\
d_y&=e_n-e_y\label{formula13}
\end{align}
where $e_n$ and $e_y$ are expression vectors of neutral and target non-neutral emotion respectively, $k$ represents the intensity, and $d_y$ is the editing direction vector from neutral emotion to the target non-neutral emotion $y$.

\textbf{ReferenceNet.} Previous work \cite{hu2024animate} shows that leveraging similar structures is beneficial in maintaining consistency in target identity. Thus we devise a UNet with a similar structure to the backbone network termed ReferenceNet, that is used to extract image features of identity frames. Specifically, the ReferenceNet is the same as the backbone network except that it does not utilize conditional residual blocks to introduce condition information such as time, audio, and expressions. Given that both ReferenceNet and the backbone network originate from the same UNet architecture, the feature maps generated by these two structures at specific layers exhibit similarities, facilitating feature maps from the two structures to fusion. In our case, the feature map of each layer extracted by the ReferenceNet is integrated into the backbone network through cross-attention. Furthermore, after performing cross-attention calculations, we introduce spatial attention to force the model to focus on important regions in the feature maps, thereby more effectively extracting useful features.

\textbf{Training Loss.} In addition to using Equations \ref{formula6} and \ref{formula7} as training losses, work \cite{stypulkowski2024diffused} also exploits lip loss to direct the model's attention to the mouth region. In our case, we also focus on the eyes and surrounding areas, as they are often closely associated with changes in emotional expression. Additionally, we use an additional eye loss to force the model to pay more attention to the eyes and surrounding region. Specifically, we use facial landmarks to locate the position of the mouth and eyes, and during training, we minimize the distance loss between the noise added to these regions and the predicted noise. With the utilization of spatial attention mentioned above, the model can effectively synthesize videos synchronized with given audio and expression conditions. Therefore, the final loss function can be expressed as:
\begin{align}
\mathcal{L}_\text{final}=\mathcal{L}_\text{simple}+\lambda_\text{vlb}\mathcal{L}_\text{vlb}+\lambda_\text{lip}\mathcal{L}_\text{lip}+\lambda_\text{eye}\mathcal{L}_\text{eye}\label{formula16}
\end{align}
where $\mathcal{L}_\text{lip}$ and $\mathcal{L}_\text{eye}$ are the noise prediction losses for the mouth and eyes region, respectively. And $\mathcal{L}_\text{simple}$ and $\mathcal{L}_\text{vlb}$ are defined by Equations \ref{formula6} and \ref{formula7}, respectively.

\subsection{Emotion Editing Condition Generation} To generate videos that reflect specific emotions during the inference phase, we need to provide emotional information to the model. Therefore, we use the DECA method to regress the expression vector corresponding to each type of emotional video. Specifically, for a video with an emotion label $y$ and length $ N$, we can establish the correspondence between emotion and expression coefficients as $\{E_y,y\}$, where $E_y=\{e^{(1)}_y,e^{(2)}_y,...,e^{(N)}_y\}$. To convert a given emotion $y$ into expression vectors $E_y$, we design a conditional LSTM-based GAN for expression vector generation, which is utilized during the inference phase of talking head video synthesis.

\subsubsection{Expression Generator} Our expression generator $G$ is designed based on LSTM architecture. Given an input emotion condition $y$, the generator $G$ synthesizes the expression vector for the current frame by utilizing the output from the previous frame, i.e., $\hat{e}^{(i)}_y=G(\hat{e}^{(i-1)}_y, y, z)$, eventually forming an expression vector sequence $\hat{E}_y=\{\hat{e}^{(1)}_y,\hat{e}^{(2)}_y,...,\hat{e}^{(N)}_y\}$. Where $z$ represents a noise vector introduced to enhance the randomness of the generated samples.

\begin{figure*}[htbp]
\centering
\includegraphics[scale=0.9]{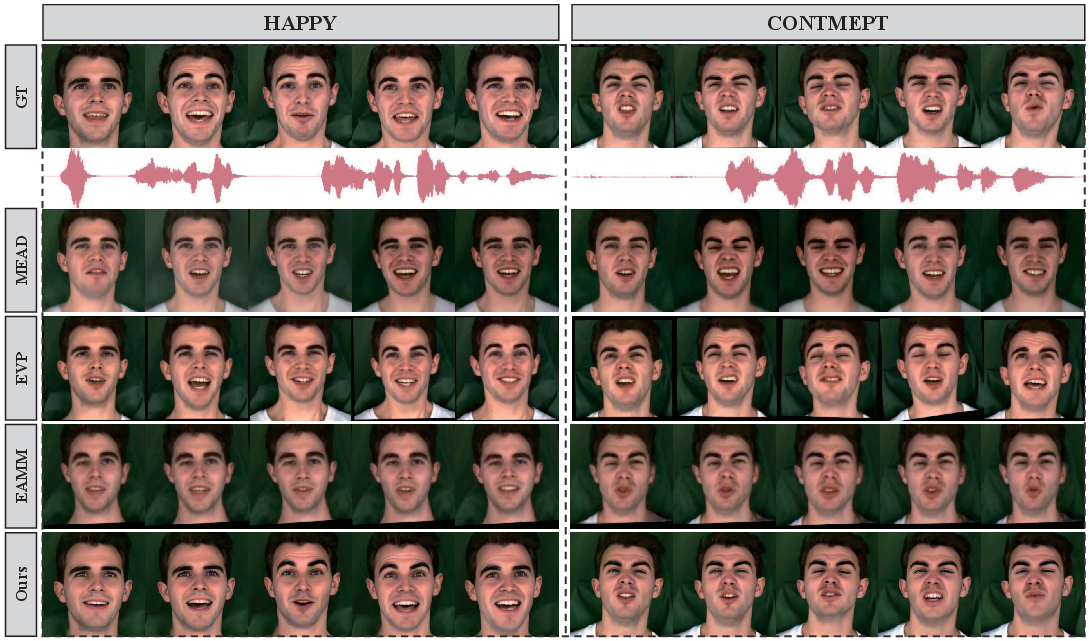}
\centering
\caption{The comparison of our model and other state-of-the-art models for emotional talking face generation. The emotion intensity value for all methods is set to 1 or the strongest.}\label{comparison}
\end{figure*}

\subsubsection{Expression Discriminator} To improve the quality of the expression vectors generated by the generator $G$, we introduce dual discriminators, $D_g$ and $D_l$, for global and local discrimination. Specifically, the discriminator $D_g$ evaluates the entire expression vector sequence to differentiate between the real sequence $E_y$ and the generated sequence $\hat{E}_y$, while also predicting its corresponding emotion label $y$. It employs a temporal convolutional network (TCN) to capture long-term dependencies and global features across the sequence, ensuring that the overall pattern and change trend of the generated expression vector sequence closely resembles those of the real samples. Meanwhile, the discriminator $D_l$ utilizes a simple MLP architecture to ensure that the expression vector $\hat{e}^{(i)}_y$ generated at each time step is similar to the real sample $e^{(i)}_y$.

\subsubsection{Training Loss} To induce the generator to generate an expression vector that adheres to the given emotion condition, we utilize three losses for adversarial training. For the global discriminator $D_g$:
\begin{align}
\mathcal{L}_\text{global} &= \mathbb{E}_{E_y,y \sim p_{\text{data}}(E_y,y)} \left[ \log D_g(E_y|y) \right] \nonumber \\
&+ \mathbb{E}_{z \sim p(z), y \sim p(y)} \left[ \log \left(1 - D_g(\hat{E}_y|y)\right) \right]\label{formula17}
\end{align}

The local discriminator $D_l$ is optimized by the adversarial loss as follows:
\begin{align}
\mathcal{L}_\text{local} &= \mathbb{E}_{E_y, y \sim p_{\text{data}}(E_y, y)} \left[ \sum_{i=1}^{N} \log D_{l}(e_y^{(i)} | y) \right] \nonumber \\
&+ \mathbb{E}_{z \sim p(z), y \sim p(y)} \left[ \sum_{i=1}^{N} \log (1 - D_{l}(\hat{e}_y^{(i)} | y)) \right]\label{formula18}
\end{align}

For the generator $G$, in addition to using the adversarial loss corresponding to $D_g$ and $D_l$, the mean square error is also introduced to minimize the distance between the generated samples and the real samples. The final loss is as follows:
\begin{align}
\mathcal{L}_G = & \, \mathbb{E}_{z \sim p(z), y \sim p(y)} [\log (1 - D(\hat{E}_y| y))] \nonumber \\
& + \mathbb{E}_{z \sim p(z), y \sim p(y)} \left[ \sum_{i=1}^{N} \log (1 - D_{s}(\hat{e}_y^{(i)}| y)) \right] \nonumber \\
& + \lambda_{\text{MSE}} \mathbb{E}_{E_y, y \sim p_{\text{data}}(E_y, y), z \sim p(z)} \| \hat{E}_y - E_y\|^2\label{formula19}
\end{align}

\renewcommand{\arraystretch}{1.15}

\begin{table*}[h]
\small
\centering
\begin{tabular}{|>{\centering\arraybackslash}m{2.5cm}| >{\centering\arraybackslash}m{1.6cm} >{\centering\arraybackslash}m{1.2cm} >{\centering\arraybackslash}m{1.2cm} >{\centering\arraybackslash}m{1.2cm}| >{\centering\arraybackslash}m{1.1cm} >{\centering\arraybackslash}m{1.1cm} >{\centering\arraybackslash}m{1.1cm}|}
\hline
Metric\textbackslash Method & EmoSpeaker \cite{feng2024emospeaker} & EVP \cite{ji2021audio} & MEAD \cite{wang2020mead} & Ground Truth & \multicolumn{3}{c|}{EMOdiffhead (Our Method)} \\  \hline
EmoAcc \cite{pham2021facial} $\uparrow$ & 0.262 & 0.438 & 0.407 & 0.457 & \multicolumn{3}{c|}{\textbf{0.477}} \\
LIE \cite{sun2023continuously} $\downarrow$ & 0.277/15 & 0.129/6 & 0.154/3 & 0.113/3 & \textbf{0.140/15} & \textbf{0.117/6} & \textbf{0.102/3}\\
FLIE $\downarrow$ & 39.553/15 & 4.016/6 & 2.778/3 & 3.116/3 & \textbf{23.406/15} & \textbf{3.804/6} & \textbf{2.104/3}\\ \hline
\end{tabular}

\caption{Comparison with other methods. In the LIE and FLIE score calculation, please note that the $A$ in $A/B$ represents the evaluation score at the $B$ emotion intensity level. All methods are tested using the results with the highest intensity level in the EmoAcc test.}
\label{tab1}
\end{table*}

\begin{table*}[h]
\small
\centering
\begin{tabular}{|>{\centering\arraybackslash}m{2.5cm} |>{\centering\arraybackslash}m{1.1cm} >{\centering\arraybackslash}m{1.1cm} >{\centering\arraybackslash}m{1.1cm} >{\centering\arraybackslash}m{1.1cm} >{\centering\arraybackslash}m{1.1cm} >{\centering\arraybackslash}m{1.1cm}|>{\centering\arraybackslash}m{1.9cm}|}
\hline
\multirow{2}{*}{Method\textbackslash Metric}  &\multicolumn{6}{c|}{Video Quality} &  Lip Sync \\ \cline{2-8}

 & FID$\downarrow$ & FVD$\downarrow$ & PSNR$\uparrow$ & SSIM$\uparrow$ &LPIPS$\downarrow$ &CPBD$\uparrow$ &$Sync_{conf}$ $\uparrow$ \\  \hline

Ground Truth  & 0.000 & 0.000 & 100.000 & 1.000 & 0.000 & 0.229 &6.493 \\

MEAD \cite{wang2020mead}  & 60.100 & 619.104 & 17.309 & 0.646 & 0.295 & 0.211 & 1.085\\

EVP \cite{ji2021audio}  & \textbf{32.278} & 225.480 &17.160 & 0.658 & 0.272 & 0.135 & 3.936\\

EAMM \cite{ji2022eamm}  & 110.068 & 873.746 & 14.417 &0.558 & 0.436 & 0.174 & 1.162\\ \hline

w/o ReferenceNet  & 68.084 & 407.488 &16.841 &0.578 &0.385 & \textbf{0.307} &4.138\\

Ours  & \underline{34.517} & \textbf{193.587} &\textbf{19.964} &\textbf{0.693} & \textbf{0.266} & \underline{0.272} &\textbf{4.722}\\ \hline
\end{tabular}

\caption{Quantitative comparisons on video quality and audio-video synchronization with other state-of-the-art methods.}
\label{tab2}
\end{table*}

\subsection{Model Inference}
For inference of the proposed method, only a driving audio, an identity image, and an emotion label with intensity are required. When an emotion label with specified intensity is given, the expression generator $G$ synthesizes the expression vector condition according to Equation \ref{formula12} for the denoising model. In the initial inference phase, we duplicate the identity image to serve as the motion frame for the denoising network. During each subsequent time step of the denoising process, the newly synthesized target frame is used as the motion frame to guide the synthesis of the next frame. Ultimately, these synthesized video frames are concatenated and passed through a super-resolution module to produce the final output talking head video.

\section{Experiment}
We train and evaluate our EMOdiffhead method on the MEAD dataset \cite{wang2020mead}, an audio-visual emotion dataset containing 60 actors across 8 different emotion categories. To capture richer facial and background information, we also incorporate the HDTF dataset \cite{zhang2021flow}, which lacks emotion labels, and train our model on both datasets. Finally, we compare our proposed method with state-of-the-art talking head video synthesis methods that support emotion editing.

\subsection{Implementation Details}
The study consists of two experimental phases, training, and inference, conducted on a computing device equipped with four 3090 GPUs. The training procedure is divided into two parts, namely the training of the expression generator and the denoising network. For the expression generator, we select videos with the highest emotion intensity from the MEAD dataset for face reconstruction to extract the expression vectors used in training. For the denoising network, to improve training efficiency, we randomly select one video for each character in the MEAD dataset under each emotion category and intensity level, while also incorporating the HDTF dataset for training. The remaining data, not used in training, is reserved for evaluating the model's generalization performance. In our case, the denoising network employs the same architecture as the UNet proposed in \cite{dhariwal2021diffusion}. The network input is images with size (128x128), that are downsampled twice to feature maps with size (64x64) and (32x32) respectively. Then, after skip connection and upsampling operations, a target frame of the same size as the input is obtained. To reduce computational costs, we only perform cross-attention calculations between the ReferenceNet and the backbone network when the feature map size is 64 and 32.

\subsection{Dataset Preprocessing}
The training videos are sampled at 30 frames per second, with the audio preprocessed to 16kHz. To enhance the quality of the synthesized video, the same face alignment is exploited in all training videos. Specifically, videos are aligned by centering on the nose tip and resized to a uniform resolution of 128x128 pixels. To compute $\mathcal{L}_\text{lip}$ and $\mathcal{L}_\text{eye}$ mentioned above, the FAN method \cite{bulat2017far} is utilized to obtain 68 facial landmarks for each frame.

\subsection{Evaluation Metric}
We employ the following metrics to evaluate the proposed method’s performance in terms of emotion linear editing, video quality, and lip synchronization.

\begin{figure}[htbp]
\centering
\includegraphics[scale=0.72]{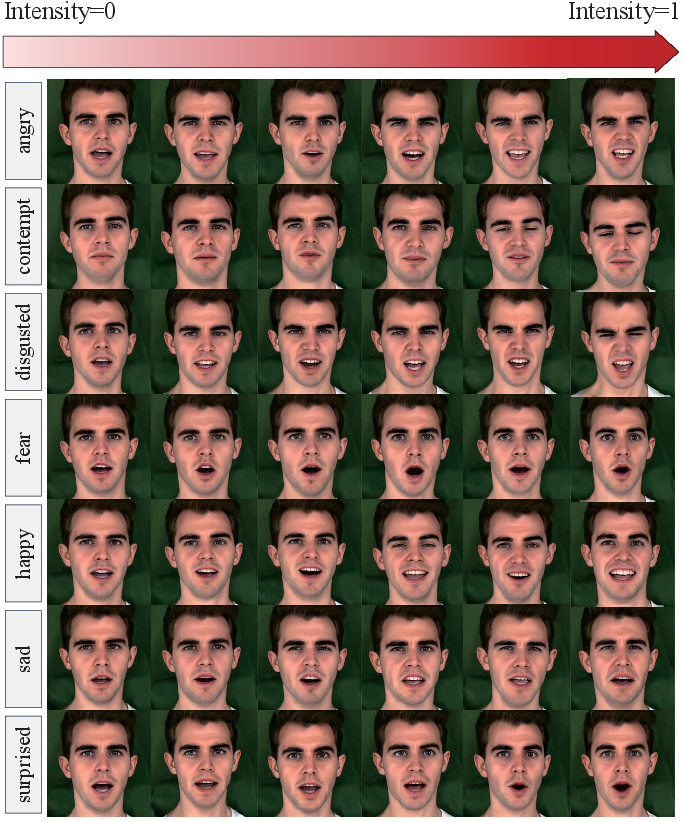}
\centering
\caption{Editing results of different facial expression types and intensities generated by our method.}\label{fine-gained}
\end{figure}

\textbf{Emotion Editing Capability.} We exploit the emotion classifier network \cite{pham2021facial} to evaluate the emotion accuracy (EmoAcc) of the generated videos. Moreover, linearity of intensity editing (LIE) \cite{sun2023continuously} is introduced to assess the linearity of EMOdiffhead's emotion editing capabilities. LIE is calculated using the pairwise perceptual difference between adjacent images with gradually changing facial expressions, based on the LPIPS model. However, the facial features extracted by LPIPS span multiple levels, from low-level visual elements to high-level semantic information, covering a broad spectrum of facial characteristics. Thus using LPIPS to evaluate the linear editing of expressions has certain limitations. To address the issue, we propose a novel metric called Flame-based linearity of intensity editing (FLIE). Given a neutral emotion vector and a target non-neutral emotion vector, our model generates a series of talking face videos where the intensity of the expressions increases linearly from 0 to 1. The DECA method is then employed to extract the expression vectors from all video frames. The difference between the average expression vectors of adjacent intensities is subsequently used to compute FLIE, which is defined as the sum of the coefficients of variation as follows:

\begin{align}
\text{FLIE}=\sum_{i=1}^{n}CV(d_{i,j})=\sum_{i=1}^{n}\frac{\sigma(d_{i,j})}{\mu(d_{i,j})},i\in(0,1],j=1,2,...,50
\label{formula20}
\end{align}
where $d_{i,j}$ represents the difference in the $j$th dimension between the average expression vector of the $i$th intensity video and the previous intensity video, $\sigma(d_{i,j})$ and $\mu(d_{i,j})$ are the standard deviation and mean of $d_{i,j}$ respectively. When the expression vector changes linearly, the difference between the expression vectors of adjacent intensity videos is almost the same, and the value of FLIE will be close to 0.

\textbf{Video Quality.} For the visual quality of synthesized faces, we use Fréchet Inception Distance (FID) \cite{heusel2017gans}, Fréchet Video Distance (FVD) \cite{unterthiner2019fvd}, structural similarity (SSIM), Cumulative Probability Blur Detection (CPBD) \cite{narvekar2011no}, Learned Perceptual Image Patch Similarity (LPIPS) and peak signal-to-noise ratio (PSNR) to analyze the generated results quantitatively.

\textbf{Audio-Visual Synchronization.} We employ the SyncNet \cite{chung2017out} to estimate the audio-visual synchronization of the synthesized results.

\subsection{Quantitative Comparison}
In this section, we quantitatively analyze the emotion editing capabilities of the proposed method and present some editing results for different emotion types and intensities shown in Figure \ref{fine-gained}. To evaluate our method, we compare it with several state-of-the-art methods, i.e., MEAD \cite{wang2020mead}, EVP \cite{ji2021audio}, and EmoSpeaker \cite{feng2024emospeaker} with emotion editing ability. Since the intensity levels of emotional face videos synthesized by different methods vary. For instance, MEAD supports only 3 intensity levels, whereas EMOspeaker allows for continuous intensity adjustment. Thus we set different intensity levels in our output videos to ensure fairness when compared with other methods. The results of this comparison are presented in Table \ref{tab1}, demonstrating that our method outperforms most existing approaches in terms of EmoAcc and linearity of intensity editing.

We also conduct quantitative comparisons on video quality and audio-video synchronization with other state-of-the-art methods \cite{ji2022eamm,wang2020mead,ji2021audio}. The results are presented in Figure \ref{comparison} and Table \ref{tab2}. As shown in Table \ref{tab2}, our proposed method performs well in terms of overall video quality and lip synchronization, and only the FID score is slightly inferior to EVP. The EVP method first generates facial landmarks and then projects them onto the edge image of the target video through 3D reconstruction. Finally, the synthesized video is obtained through rendering. Therefore, it achieves a head posture that more closely resembles that of the target video. Additionally, we observed that the faces generated by EAMM differ significantly from the reference image, and emotion control relies heavily on the reference video, making it extremely inconvenient. MEAD and EVP also lack accuracy in lip movements and are unable to produce emotionally expressive videos with fine-grained control. In contrast, our proposed method ensures a high degree of lip synchronization and richer emotional expression while maintaining video quality.

\begin{figure}[htbp]
\centering
\includegraphics[scale=0.43]{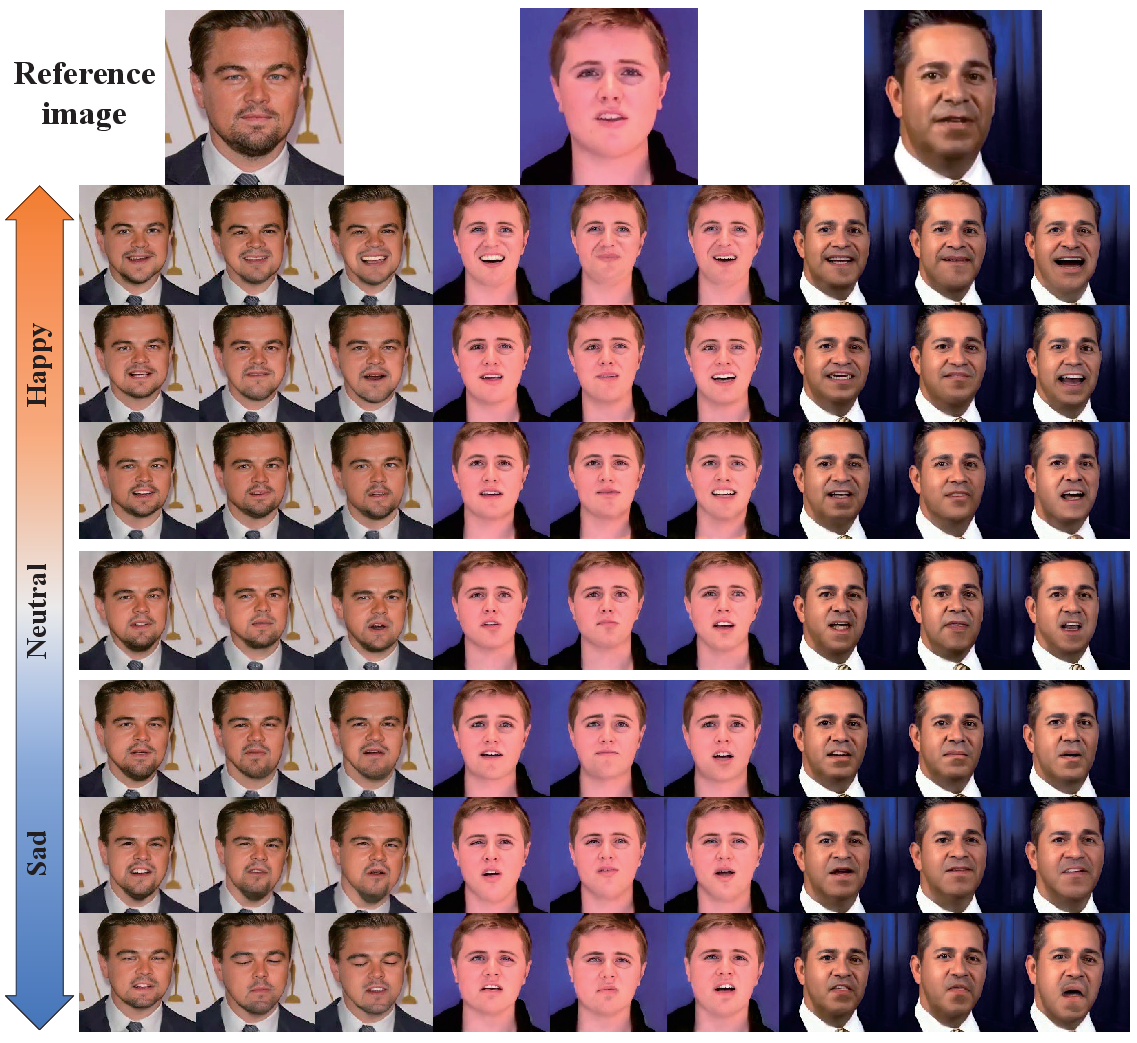}
\centering
\caption{One-shot emotion generation results on images not seen during training. The direction of the arrow represents the increasing intensity of the emotion.}\label{oneshot}
\end{figure}

Moreover, unlike EVP and MEAD, our method can generalize to other identity images. To verify the stability of the model, we conduct one-shot generation on images from the HDTF dataset that are not seen during training and do not contain any emotional information, as shown in Figure \ref{oneshot}. The results show that our method successfully controls the emotion category and intensity in unseen images.

\subsection{User Study}
We conduct a user study to compare EMOdiffhead with other SOTA methods. Specifically, 20 participants are invited to assess four methods' results and ground truth videos in terms of emotion accuracy, audio-visual synchronization, identity consistency, and video quality. When participants are presented with 5 videos from different methods (including ground truth video), they are required to rate the above four criteria on a scale of 1 to 5. The obtained scores are averaged to get the final result reported in Table \ref{tab3}. The results indicate that our method outperforms other state-of-the-art methods in terms of emotion accuracy, video quality, and audio-video synchronization. Only in identity consistency, the score is inferior to EVP and MEAD, because both of them are trained on only a few identity data and thus maintain good performance in generating identity-specific videos.

\begin{table}[h]
\small
\centering
\begin{tabular}{|>{\centering\arraybackslash}m{1.85cm}| >{\centering\arraybackslash}m{1.2cm} >{\centering\arraybackslash}m{0.75cm} >{\centering\arraybackslash}m{1.4cm} >{\centering\arraybackslash}m{1.1cm}|}
\hline
Method\textbackslash Metric & Emotion Accuracy$\uparrow$  & Lip Sync$\uparrow$ & Identity Consistency$\uparrow$ & Video Quality$\uparrow$ \\  \hline
Ground Truth &4.86 &4.81 &4.74 &4.54 \\

MEAD &3.99 & 2.44 &4.09 &2.88 \\

EVP  &4.03 &3.21 &\textbf{4.25} &3.85 \\

EAMM &2.18 &3.15 &1.73 &2.19\\ \hline
Ours  &\textbf{4.50} &\textbf{4.41} &3.83 &\textbf{4.18}\\ \hline
\end{tabular}

\caption{Results of a user study on 5 types of videos from different methods (including ground truth videos)}
\label{tab3}
\end{table}

\subsection{Ablation Study}
\subsubsection{Ablation of Emotion Condition Generation} To enable the emotion generation network to synthesize emotion editing conditions of any intensity, we use the videos with the highest emotion intensity from the MEAD dataset to train the network. During inference, it is only necessary to generate the target emotion expression and the neutral emotion expression, and any desired intensity level of emotional expression can then be produced according to Equations \ref{formula12} and \ref{formula13}. In the discriminator setup, we employ TCN blocks in the global discriminator to enhance the temporal dynamics of the generated emotional conditions. Additionally, a local discriminator is introduced to improve the authenticity of the emotion vector for each frame.

\begin{figure}[htbp]
\centering
\includegraphics[scale=0.32]{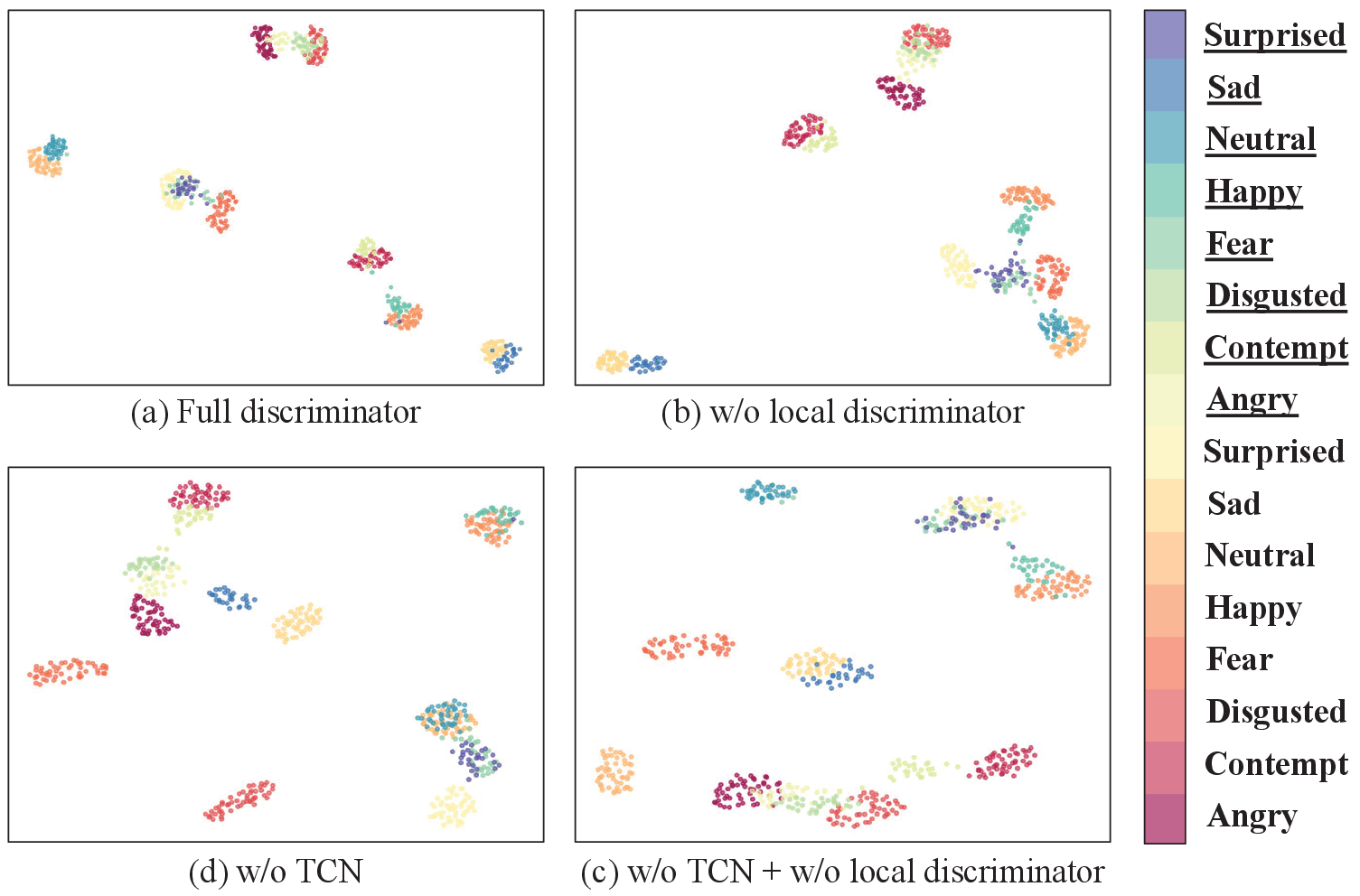}
\centering
\caption{The visualization shows the emotional condition reconstruction results under different discriminator setups: (a) TCN is applied in the global discriminator and the local discriminator is introduced. (b) The local discriminator is removed. (c) TCN in (a) is replaced with a standard convolution. (d) The local discriminator is removed based on the setup in (c). Different colors represent different emotions, with the underlined ones representing the emotional conditions generated by the generator.}\label{UMAP}
\end{figure}

To verify the effectiveness of this setup, we conducted ablation experiments on the discriminator. Specifically, we design different global discriminators, both with and without local discriminators, resulting in four types of discriminators to evaluate their performance. The UMAP \cite{mcinnes2018umap} visualization in Figure \ref{UMAP} shows that our settings significantly improve the ability to reconstruct emotion expression conditions. Furthermore, as reported in Table \ref{tab4}, our discriminator configuration achieves the best performance in conditioning the backbone network to synthesize videos that match the expression vector conditions.

\begin{table}[h]
\small
\centering
\begin{tabular}{|>{\centering\arraybackslash}m{2.8cm}| >{\centering\arraybackslash}m{1.2cm} >{\centering\arraybackslash}m{1cm} >{\centering\arraybackslash}m{1.2cm} |}
\hline
Setting\textbackslash Metric & EmoAcc$\uparrow$& LIE$\downarrow$ & FLIE$\downarrow$\\  \hline
full disc&\textbf{0.477} &0.140 &\textbf{23.406} \\

w/o local disc &0.376 &\textbf{0.124} &335.007 \\

w/o TCN  &0.468 &0.171 &70.922 \\

w/o TCN + local disc&0.369 &0.130 &147.097\\ \hline
\end{tabular}

\caption{Evaluation results of emotion accuracy, LIE, and FLIE of synthesized videos under different discriminator setups. When calculating FLIE and LIE scores, we set the intensity level to 15.}
\label{tab4}
\end{table}

\subsubsection{Ablation of ReferenceNet} To assess the effectiveness of the ReferenceNet in maintaining identity consistency, we evaluate the quality of synthesized videos using a denoising network without the ReferenceNet. The results as shown in Table \ref{tab2}, indicate that the denoising network with the ReferenceNet generally outperforms the one without it in terms of video quality. It demonstrates that incorporating the ReferenceNet contributes to preserve the target identity, thereby enhancing video quality.

\section{Conclusion}\label{sec.conclusion}
In this paper, we present EMOdiffhead, a novel method for emotion-driven talking head video generation that provides fine-grained control over both emotion type and intensity. Given the linearity of the FLAME model in facial expression modeling, we employ the DECA method to extract expression vectors, which guide the diffusion model in generating videos synchronized with emotion. EMOdiffhead effectively overcomes the limitations of emotional data, such as the lack of diversity in facial and background information, and addresses the absence of emotional details in emotion-irrelevant data. It not only provides fine-grained control over facial expressions but also supports one-shot generation. Extensive experiments demonstrate that our method achieves state-of-the-art performance, offering significant advancements in the field of emotion-driven portrait animation.

{\small
\bibliographystyle{ieee_fullname}
\bibliography{egbib}
}
\end{document}